*Article*

# A genetic algorithm for structure-activity relationships: software implementation

**Lorentz JÄNTSCHI**

Technical University of Cluj-Napoca, 400641 Cluj, Romania
E-mail: lori@academicdirect.org

**Abstract:** The design and the implementation of a genetic algorithm are described. The applicability domain is on structure-activity relationships expressed as multiple linear regressions and predictor variables are from families of structure-based molecular descriptors. An experiment to compare different selection and survival strategies was designed and realized. The genetic algorithm was run using the designed experiment on a set of 206 polychlorinated biphenyls searching on structure-activity relationships having known the measured octanol-water partition coefficients and a family of molecular descriptors. The experiment shows that different selection and survival strategies create different partitions on the entire population of all possible genotypes.

## 1. Introduction

First simulations of evolution were found in studies of Nils Aall BARRICELLI [1-4]. Shortly later, Alex FRASER (1923-2002) it published a series of papers about simulation of artificial selection of organisms with a measurable trait loci. Fraser and collaborators simulations (Fraser, 1957-1970) include all essential elements of modern genetic algorithms [5-19].

Even most of the heuristics are in significant measure ad-hoc and dependent to the given problem, the developing of the informatics lead scientists to formulate three heuristics which are very general and applicable to a large variety of hard problems; due to their generality they now are called meta-heuristics. All three are stochastic in nature, being based on occurring natural processes, and together with genetic algorithms (GA), of which great expansion of studies started around 1970's [20,21] and reinvented later [22,23], this family also contains simulated annealing (SA) [24,25] and tabu search (TS) [26-28].

The quality of an heuristic algorithm is defined through three criteria: it's speed (how fast it obtain the solution), it's precision (how far is the solution to the global optimum), and it's scope or applicability domain (how large is the subset of the input data relative to the entire set of possible values for which the algorithm performs according to previous two criteria). An important issue is of algorithmic complexity appears here, and is sustained by the No Free Lunch Theorem (NFLT) [29,30]. The NFLT theorem shows that by using all three criteria, all algorithms are equivalent (being A and B algorithms, and being a dataset for which A performs better than B, it exists a dataset for which B performs better than A). The important consequence is about the design methodology: in order to construct a good algorithm, the key is the well-defining of the applicability domain, for which a given algorithm which we intend to create, may perform better than others.

The genetic algorithms supposes adaptive heuristic search being based on the ideas of the evolution; thus they bring the concepts of natural selection and genetics into the mathematical simulation by the uses of the computer. Mimics of the observed processes in natural evolution of the organic matter it serves as tool for solving decision, classification, optimization, and simulation problems. The key elements which are called to contribute in a genetic algorithm are: the genetic model (the phenotype-genotype dualism) such as was formulated and augmented from first steps of modern genetics [31,32]; crossover (the traits - genes dualism) such as it was observed by the precursors of modern genetics [33-35]; mutation, such as was observed beginning with modern genetics precursors to date (random [36], deliberate through exposing to certain conditions [37,38], or under environment factors stress [39]); and finally, the last but not the list, natural selection and survival of the fittest [40].

The obtaining of a good structure-activity relationship is a hard problem containing through its complexity all ingredients for a model of a hard-problem. Thus, characterizing of the relationship between structure and biochemical activity implies all categories of hard-problems: optimization (of the structure-activity model by maximizing its capacity of estimation and prediction), classification (use of the model to classify the compounds into classes of activity), decision (use of the model to take a decision regarding the synthesising of a new compound for which the model predicts better activity).

The hard problem of structure-activity relationship is as follows: having a structural information (obtained from molecular topology and geometry) and a biochemical information (obtained from a designed experiment), which is the best structure-activity relationships describing the activity (biochemical information) depending on structure (structural information).

A suitable way for genetic algorithms applicability domain is to construct a family of structural descriptors, such as Molecular Descriptors Family (MDF), described in [41].

The aim of this research is to define the frame of the implementation of a genetic algorithm of which applicability domain is on structure-activity relationships on families of molecular descriptors.

## 2. Material and Method

The work [42] may serve as example of a hard problem of biochemical structure vs. measured property. Here, starting from experimental data of observed retention times of polychlorinated biphenyls (PCBs) reported in [43], a database containing structure based information was build; figure 1 depicts the procedure of obtaining the information.

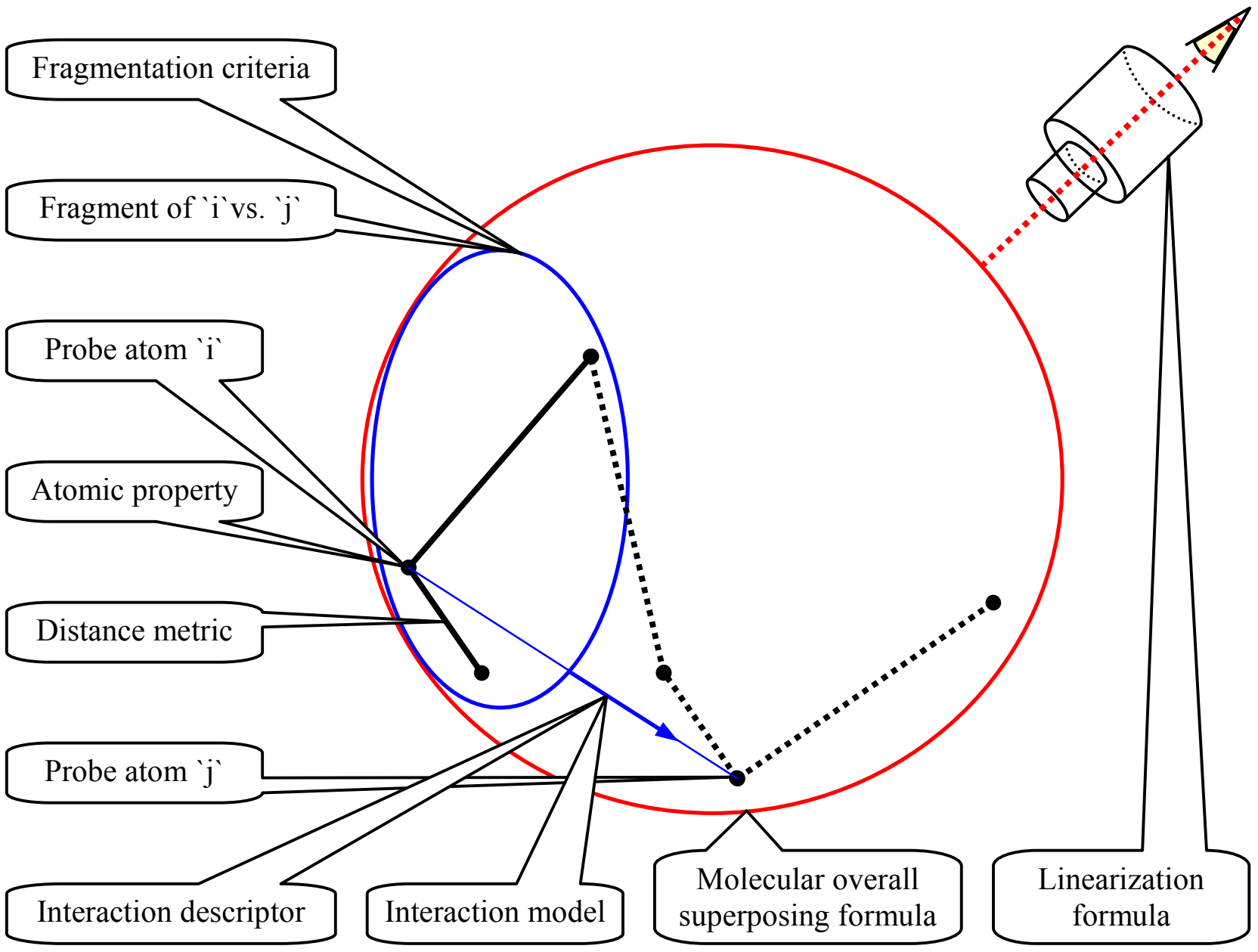

**Figure 1.** Methodology of structure-activity relationships (SAR) in Molecular Descriptors Family (MDF) approach (image from [44])

In order to a genetic algorithm to become suitable for structure-activity relationships, we must define its search space. Thus, the problem of structure-activity relationship finding must be formulated using genetic terms.

Every gene (one of the values from Gene columns of Table 1; ex. $I_M$ for FPIF family of structural descriptors) encodes an operator used to construct the chromosome (gene sequence of a family in Table 1; ex. $D_MA_PI_DI_MF_CS_ML_O$ for MDF) of a molecular descriptor (Table 1). Every descriptor of a family of descriptors is a genotype (a possible series of values for every gene of a chromosome; ex. TCJtAAfDI for MDFV) and all together constitutes the genetic material (the set of all possible combinations of values from Genome column in Table 1; ex. {R, D} × {T, G} × {M, E, C, Q} × {__p__, __d__, _1/p_, _1/d_, _p*d_, _p/d_, _p/d2, p2/d2} × {si, se, ji, je, fi, fe} × {S, P, A, G, H} ×{P_, P2, E_, E2} × {I, R, L} × {t, g} for FPIF) of the given family.

The number of encoded values of the genes varies from two values (ex. the gene encoding the metric type - topological of geometrical distance - $D_M$ for FPIF and MDF; $D_O$ for MDFV) to fifty-eight (the $I_D$ interaction descriptor of MDFV family). The volume (size) of the genetic material varies from a family to another and Table 2 summarizes these variations.

**Table 1.** Search space of three families of molecular descriptors

| Family | Gene | Genome |
|---|---|---|
| FPIF [45] | $I_M$ | R D |
| | $D_M$ | T G |
| | $A_P$ | M E C Q |
| | $P_D$ | _p_ _d_ _1/p_ _1/d_ _p*d_ _p/d_ _p/d2 p2/d2 |
| | $F_C$ | si se ji je fi fe |
| | $S_M$ | S P A G H |
| | $M_I$ | P_ P2 E_ E2 |
| | $L_O$ | I R L |
| MDF [46] | $D_M$ | t g |
| | $A_P$ | C H M E G Q |
| | $I_D$ | D d O o P p Q q J j K k L l V E W w F f S s T t |
| | $I_M$ | r R m M d D |
| | $F_C$ | m M D P |
| | $S_M$ | m M n N S A a B b P G g F f s H h I i |
| | $L_O$ | I i A a L l |
| MDFV [47] | $D_O$ | T G |
| | $A_P$ | C H M E Q L A |
| | $I_D$ | J j O o P p Q q R r K k L l M m N n W w X x Y y Z z S s T |
| | | t U u V v F f G g H h I i A a B b C c D d 0 1 2 3 4 5 6 7 |
| | $S_F$ | A a I i F P C |
| | $S_M$ | A a I i F P C |
| | $I_T$ | f F c C p P a A i I |
| | $E_U$ | D d |
| | $L_O$ | I R L |

**Table 2.** Volumes of the molecular descriptors families from Table 1

| Family | Gene | | | | | | | | Volume (N) |
|---|---|---|---|---|---|---|---|---|---|
| FPIF ([45]) | $I_M$:2 | $D_M$:2 | $A_P$:4 | $P_D$:8 | $F_C$:6 | $S_M$:5 | $M_I$:4 | $L_O$:3 | 46080 |
| MDF ([46]) | $D_M$:2 | $A_P$:6 | $I_D$:6 | $I_M$:24 | $F_C$:4 | $S_M$:19 | $L_O$:6 | | 787968 |
| MDFV ([47]) | $D_O$:2 | $A_P$:7 | $I_D$:58 | $S_F$:7 | $S_M$:7 | $I_T$:10 | $E_U$:2 | $L_O$:3 | 2387280 |

Searching on the molecular descriptors space is done with multiple linear regressions of type (1a) or type (1b), where Y is the array of the activity experimental measurements (the dependent variable; under assumption of experimental random error), $\{X_1, \ldots, X_n\}$ a set of descriptors drawn from a family (independent variables; under assumption of linear association with observed Y), and $(b_i)_{i\leq n}$ are the model parameters (to be obtained under assumption of least squares error):

$$b_0 + b_1X_1 + \ldots + b_nX_n = \hat{Y} \sim Y \quad (1a)$$

$$b_1X_1 + \ldots + b_nX_n = \hat{Y} \sim Y \quad (1b)$$

Following characterizes the equations (1): n (or |X|) - the number of independent variables; m (or $|Y| = |X_1| = \ldots = |X_n|$) - the number of experimental observations; |b| (n+1 for (1a) and n for (1b)) - the number of unknown parameters of multiple linear regression model (1a) or (1b).

Following assumptions are made: the values of Y are normal distributed (thus, as example, Figure 2 depicts frequency distribution of the measured octanol/water partition coefficients of a series of 206

PCBs from a total of 209 expressed in logarithmic scale, as is entered into analysis in [48]); the measurement error of Y is random and normal distributed too; the $X_1$, …, $X_n$ variables are normal distributed and no affected by errors.

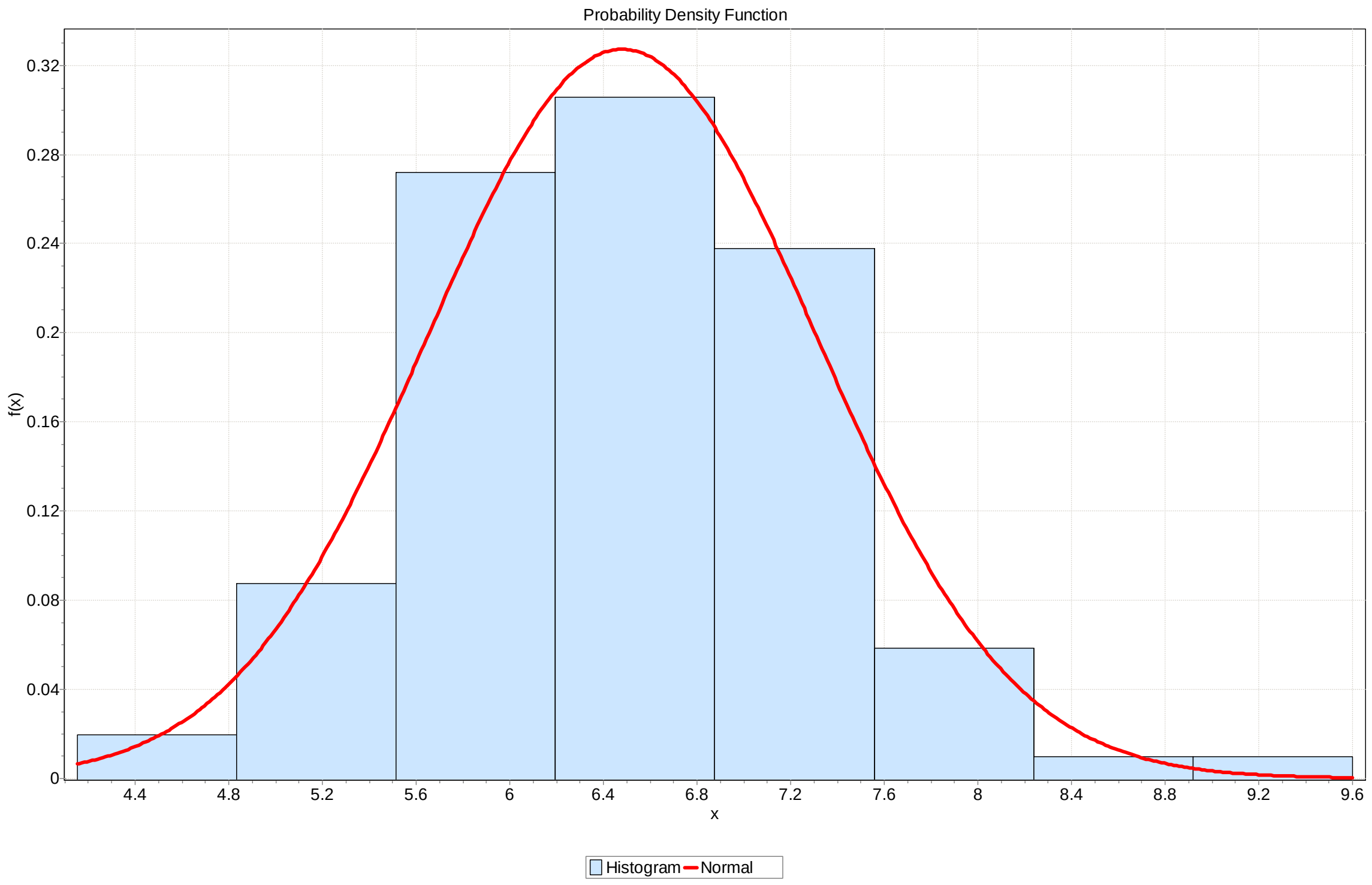

**Figure 2.** Histogram of observed log($K_{ow}$) for 206 PCBs and maximum likelihood estimation of PCBs population distribution (mean: 6.4806; standard deviation: 0.83076)

Obtaining of the regression parameters $(b_i)_{i \le n}$ is always accompanied by a risk being in error, and under hypothesis that exists the linear relationship defined by (1a) or (1b) their statistical significance and confidence intervals may be obtained by using Student t distribution [49,50]. In order to (1a) or (1b) have unique solution is necessary (but not enough) that $|b| \le m\text{-}1$; in order to $(b_i)_{i \le n}$ have statistical significance is necessary (but not enough) that $|b| \le m\text{-}6$. If $b_0$ of (1a) has no statistical significance, then we should use equation (1b) as alternative of the more general case (1a). The no statistical significance of a coefficient $b_i$ for $1 \le i \le n$ in equation (1a) associated with the absence of its statistical significance in (1b) should lead to rejecting of the hypothesis of the existence of the linear relationship between $X_i$ and Y.

The size of the search space ($V_s$) is a function of descriptors family size (N, Table 1) and the multiplicity of the linear regression (n, eq.1):

$$V_S = \prod_{j=1}^{n} \frac{N - j + 1}{j} = \binom{N}{n} \tag{2}$$

The equation (2) allows expressing of the calculus complexity to browse the entire searching space; the value of the eq.2 may be doubled if the search are conducted by both (1a) and (1b) equations. It may be checked that eq.2 defines a hard problem (problem of thich solution obtained by the best imaginable algorithm require an execution time increasing exponentially by the size of the input data).

Representing graphically (Figure 3) the equation (2) for different values of n (number of independent variables in the expression of the multiple linear regression) it comes the proof (of exponential dependence) too.

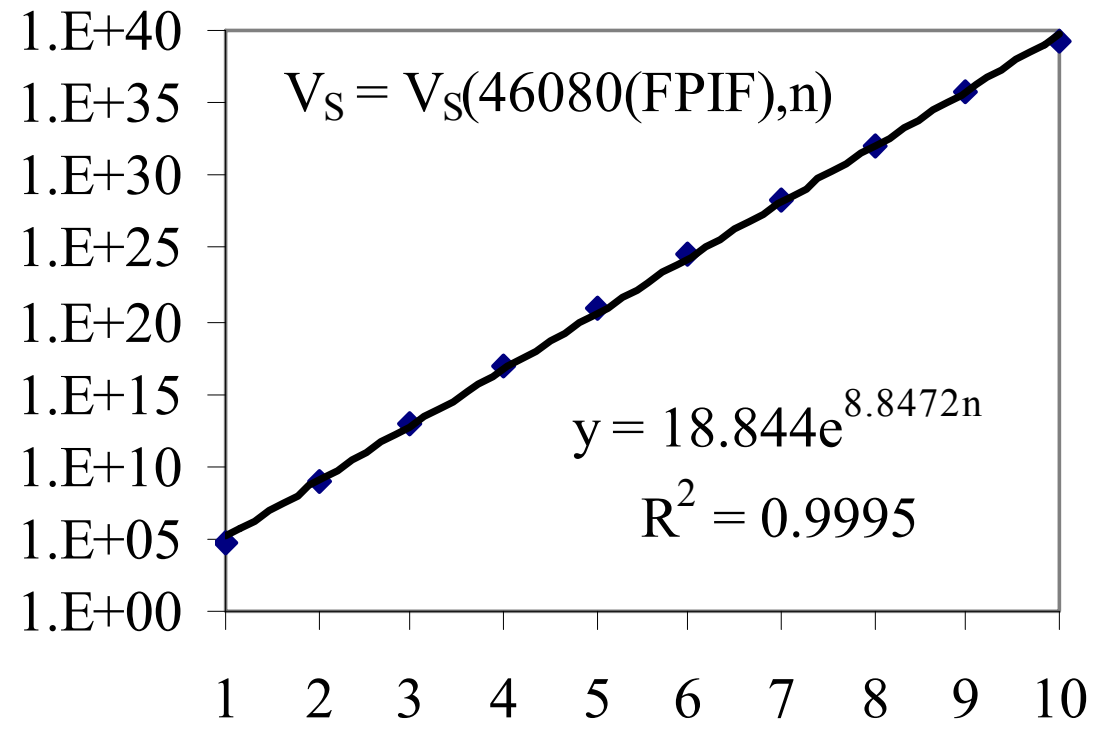

**Figure 3.** Exponential complexity of multiple linear regressions with families of molecular descriptors (example for FPIF)

## 3. Genetic Algorithm Implementation

The design of a genetic algorithm supposes initialization (of random or deterministic type) of a sample (a subset of the genetic material of the molecular descriptors family; ex. {tCDrmmI, gHdRMMi, gMddMMi} is a sample of size 3 from MDF) of chromosomes from genetic material; let be *p* the size of the sample; then $X_1$, …, $X_p$ enters into the evolution process (genetic complex process which implies selection, crossover and mutation processes) into the cultivar (seen as memory or virtual space in which the genotypes are transformed into phenotypes through applying of the operators defined by the gene values for the entire set of *m* molecules; the phenotype associated with the genotype is thus a array of *m* numerical values, one each for every molecule of the set).

The genetic algorithm (seen thus as an algorithm which describes through instructions the evolution process applied to the sample) operates on the sample which content are modified in every generation (a generation being one iteration of the genetic algorithm).

Every set of *n* distinct descriptors is a point in the search space (the set of all possible selections of *n* descriptors from a total of *N* - eq. 2) and in same time a possible solution (a regression equation of generic type defined by eqs.1). Basic operations of a genetic algorithm are chromosomes crossover and mutation (Figure 4).

***Crossover*** (the process through which a sequence of the chromosome are replaced by the corresponding sequence of another chromosome and vice versa; crossover is made hoping that by recombining of sequences of good genotypes is likely to produce even better descendants than the parents from which it comes) of two genotypes suppose the choosing (random or deterministic) of a contiguous sequence to be crossed over from the gene array; the values of the sequences are exchanged and two descendants are obtained.

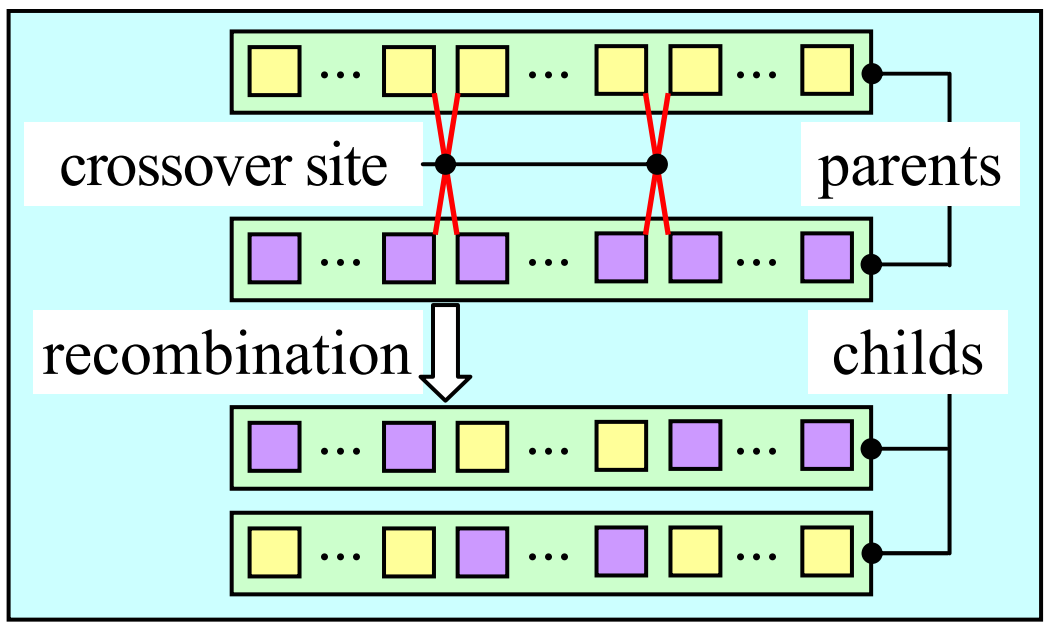

Crossover

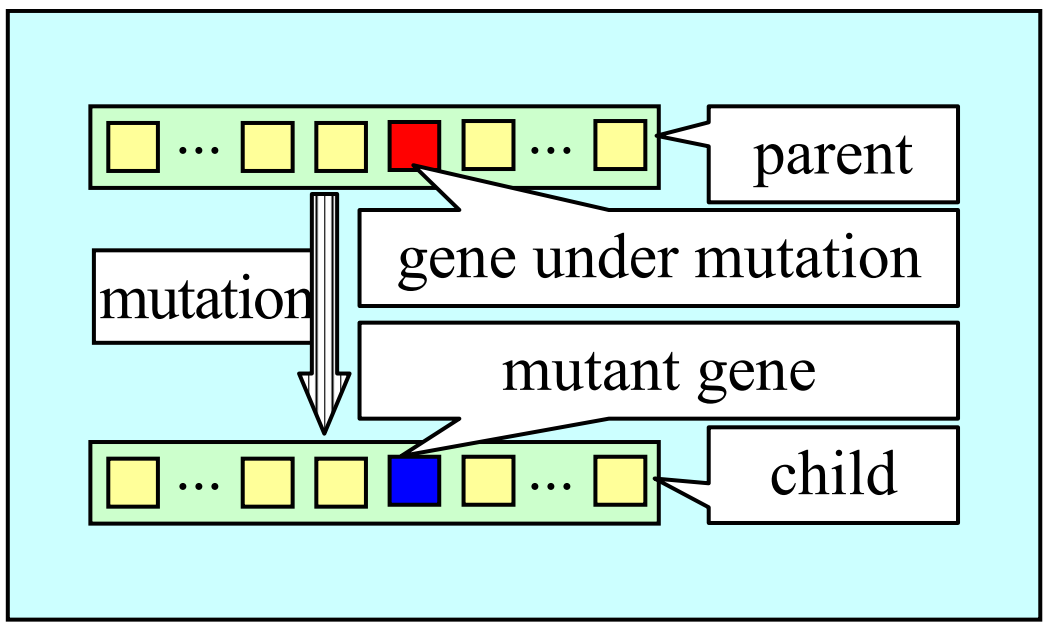

Mutation

**Figure 4.** Crossover and mutation

***Mutation*** (the operator which introduces new modifications - inexistent in the sample in the generation; a characteristic is its occurrence with low probability, being thus applied with a low probability) of a genotype suppose changing of a value of a gene of the chromosome with other value from the list of possible values for the gene. The result of the mutation and crossover are the descendants (being thus the genotypes obtained from crossover and eventually mutation of the individuals from sample) or childs.

***Selection*** (the operator with which one or more individuals are extracted from the sample in order to participate at proliferation) is the implicit operation required in order to mutate and crossover, and acts based on a selection score (a numerical value associated to the individual and calculated or expressed from the fitness of the phenotype into its cultivar environment). At least a part of the descendants should be viable descriptors (phenotypic viability - refers its potential to be used in regressions; a descriptor is viable if has real and finite values for all molecules from the dataset and not all its values are identical; supplementary further conditions can be imposed, such as: reasonable variability - by using the coefficient of variation, reasonable departure from normality - by using a normality test such as Jarque-Bera [ 51 ], and a reasonable explanation power - by using its determination coefficient from simple linear regression with measured activity), being then able to be part of candidate solutions.

Viable descendents replaces a corresponding part of the individuals from the sample through a process of ***survival*** (the operator with which one or more individuals are removed from the sample their places being taken by the descendants) applied to the individuals based on a survival score (a numerical value associated to an individual based on both genotyping - measuring its genotypic similarity with all other individuals from cultivar with the purpose of maintaining the diversity of the genetic material from cultivar) and phenotyping - measuring its phenotypic similarity with the purpose of maintaining the diversity of traits).

The last but not the list parameter of the genetic algorithm is the ***evolution objective*** (the parameter under optimization; minimization - ex. sum of squares error; maximization - ex. determination coefficient) measured by a objective function (seen as the algorithm of calculation of the trait which constitutes the objective of the sample evolution).

An option is available: too keep (and in this case being excluded from survival procedure) or not into the cultivar the individuals with the best value of the objective function.

As it results, not all individuals of a generation survive and are included into the next generation. The reason to do this is in order to keep constant the number of the individuals in cultivar (or cultivar size). Thus, the number of the replaced individuals is equal with the number of viable descendants.

Selection and survival based on the selection and survival scores is implemented via ***selection and survival strategies*** (strategy - extraction method of an individual from sample using scores). What Table 3 formally presents express the fact that three alternatives are used (proportional, deterministic, and tournament) applicable to scores or to scores ranks (when ranks replaces values); more, the value of the score can enter into a process of normalization (of which purpose is correcting - relative adjusting - of the individuals scores relative to two references - one for minimum, and one for maximum - which are global updated in every generation during the entire evolution process). Score functions ($f(i) = f_i$ in Table 3) may have different expressions from evolution objective (Table 4 - evolution objective scores) to selection (Table 5 - selection scores) and to survival (Table 6 - survival scores).

**Table 3.** Selection and survival strategies

| Method | Score function | Extraction | Comments |
|---|---|---|---|
| Proportional | $f_i$ = Fitness(Chromosome_i) | $p_i=f_i/\Sigma_i f_i$ | Likelihood proportional with the score (by using the $p_i$ probability to extract) |
| Deterministic | | $i \mid f_i$ = max. or min. | Extraction of the strongest or of the weakest individual (elitism) |
| Tournament | | $(f_i,f_j)$ max. or min. | Pairs of individual competes for extraction |
| Normalization | $g_i = (f_i-N_0)(f_{max.}-f_{min.})/(N_1-N_0)$ | | A fixed scale $[N_0,N_1]$ normalizes scores between generations |
| Ranks | $h_i = \mathrm{Rank}(f_i)(f_{max.}-f_{min.})/\mathrm{Size}$ | | Rank scores replaces scores |

**Table 4.** Objective scores for multiple linear regressions

| Score | Meaning | Objective | Remarks |
|---|---|---|---|
| $se_s(\hat{Y})$ | Sum of estimation errors | minim | Usually s=2; for s = 1 (and more for s =1/2) the general tendency of regression are more weighted in disfavour of grosser deviations from regression line |
| $r2_s(\hat{Y})$ | Determination coefficient | maxim | Usually s = 1; Most common objective (highest determination) |
| $Mt_s(t)$ | Minkowski mean of significances | maxim | Give weights to the significance of every parameter from the regression ($t_i = t(b_i)$); t - Student t statistic |
| $Hr_s(r^2)$ | Shannon entropy of determination | minim | It uses a logarithmic scale for expressing the objective (in bits) |
| $\hat{Y}=a_0+\Sigma_{1\le i\le n}a_i\cdot\mathrm{Phenotype}_i$ or $\hat{Y}=\Sigma_{1\le i\le n}a_i\cdot\mathrm{Phenotype}_i$ (when $a_0 \neq 0$ not statistically significant)<br>$se_s(\hat{Y}) = \sum_{i=1}^{m} \lvert \hat{Y}_i - Y_i \rvert^s$; $r2_s(\hat{Y}) = \left(r^2(Y,\hat{Y})\right)^s$;<br>$Mt_s(t) = \left(\frac{1}{n}\sum_{i=1}^{n} t_i^{\,s}\right)^{1/s}$; $Hr_s(r^2) = \frac{\log_2\left(r^{2s}+(1-r^2)^s\right)}{1-s}$ | | | |

**Table 5.** Selection scores for multiple linear regressions

| Score | Meaning (objectives as in Table 4) |
|---|---|
| nalive($X_i$) | number of valid regressions containing $X_i$ phenotype |
| r2_min($X_i$) | lowest $r2_s(\hat{Y})$ from all valid regressions containing $X_i$ phenotype |
| se_min($X_i$) | lowest $se_s(\hat{Y})$ from all valid regressions containing $X_i$ phenotype |
| Mt_min($X_i$) | lowest $Mt_s(t)$ from all valid regressions containing $X_i$ phenotype |
| Hr_min($X_i$) | lowest $Hr_s(r^2)$ from all valid regressions containing $X_i$ phenotype |
| r2_max($X_i$) | highest $r2_s(\hat{Y})$ from all valid regressions containing $X_i$ phenotype |
| se_max($X_i$) | highest $se_s(\hat{Y})$ from all valid regressions containing $X_i$ phenotype |
| Mt_max($X_i$) | highest $Mt_s(t)$ from all valid regressions containing $X_i$ phenotype |
| Hr_max($X_i$) | highest $Hr_s(r^2)$ from all valid regressions containing $X_i$ phenotype |
| r2_avg($X_i$) | average $r2_s(\hat{Y})$ from all valid regressions containing $X_i$ phenotype |
| se_avg($X_i$) | average $se_s(\hat{Y})$ from all valid regressions containing $X_i$ phenotype |
| Mt_avg($X_i$) | average $Mt_s(t)$ from all valid regressions containing $X_i$ phenotype |
| Hr_avg($X_i$) | average $Hr_s(r^2)$ from all valid regressions containing $X_i$ phenotype |

**Table 6.** Survival scores for multiple linear regressions

| Score | Meaning | Objective | Remarks |
|---|---|---|---|
| $VSP_q(X_i, X_j) = \vert f(X_i) - f(X_j) \vert^q$ | phenotypes dissimilarity | minimum | f(·) is the selection score |
| $VSG_r(X_i, X_j) = \left(\frac{NCD(X_i, X_j)}{NC}\right)^r$ | genotypes dissimilarity | minimum | NCD(·,·) - number of different gene values of given parameters<br>NC - number of genes in chromosome |
| $VS(X_i, X_j) = \frac{2}{VSP_q(X_i, X_j) + VSG_r(X_i, X_j)}$ | pair similarity | maximum | VS(·,·) is a measure of likelihood |
| $VS(X_i) = \min_{1 \le j \le n}^{j \ne i} VS(X_i, X_j)$ | individual similarity | maximum | Worst case defines the score |

In order to solve a hard problem of multiple linear regressions with a family of molecular descriptors the ***genetic algorithm*** (Figure 5) random (or deterministic) generates a sample of *genotypes* of a given size ($p$, kept the same during the evolution, $n < p < N$) and then repeat:

÷ (Step_1) Obtain *phenotypes* from genotypes;

÷ (Step_2) Compute $C_n^p$ *multiple linear regressions* of type (1a) and eventually (1b); output (or store) the best found model and eventually (if is configured to do so) mark as survived the phenotypes which acts as descriptors in the model; store regression scores;

÷ (Step_3) Obtain *objective scores* of the individuals from regression scores;

÷ (Step_4) Obtain *selection scores* of the individuals;

÷ (Step_5) Using selection strategy extract from sample $k$ ($k$ given) *pairs of genotypes*;

÷ (Step_6) With a low probability (*pp*) *mutate* every of $2k$ genotypes (parents);

÷ (Step_7) *Crossover* the k pairs of genotypes and obtain $2k$ new ones (descendents);

- ÷ (Step_8) With a low probability (*cp*) *mutate* every of 2*k* genotypes (childs);
- ÷ (Step_9) Obtain the viable (adapted to the environment) *childs subset* (of size $v \leq 2k$);
- ÷ (Step_10) Obtain *survival scores* of the (remained) individuals (genotyping and phenotyping);
- ÷ (Step_11) Using survival strategy remove from sample v individuals and *replace* it with childs subset;

until best found model meet the requirements or a given number of iterations are exhausted.

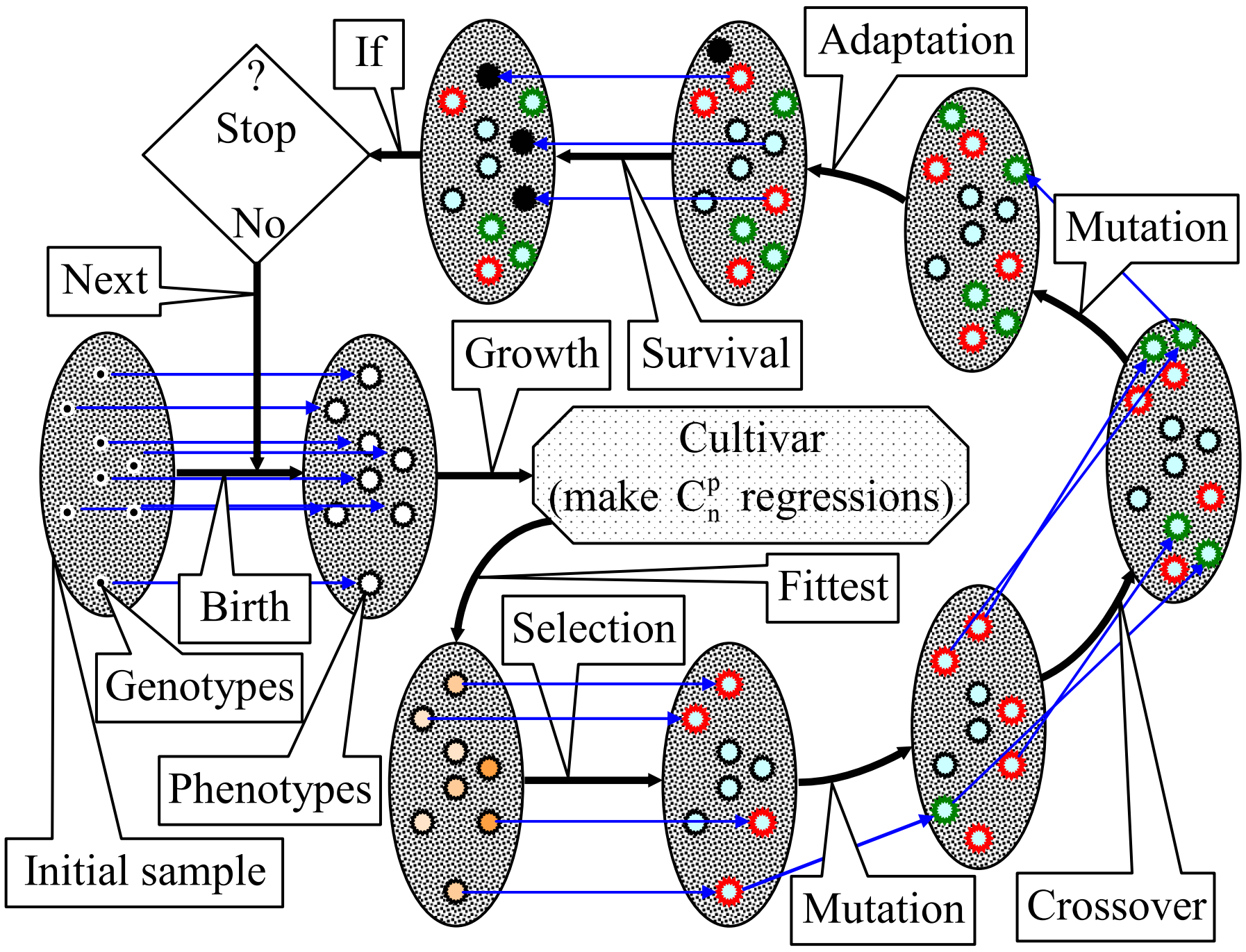

**Figure 5.** The genetic algorithm: evolution

The implementation of the genetic algorithm reveals a series of issues regarding selection and survival strategies requiring special care of, these being solved as follows:

***Selection (fittest) scores (FS)*** come from:

- ÷ Computing of the all possible regressions between phenotypes and recording of the valid ones selection scores;
- ÷ Computing of the selection scores by phenotype from all its occurrences in regressions;
- ÷ Computing of the selection scores by genotype from all its occurrences in phenotypes;
- ÷ Normalizing of the scores between generations if was configured to do;
- ÷ Rounds to the defined number of significant digits;
- ÷ Build ranks of the scores;
- ÷ Replaces scores with ranks if was configured to do;
- ÷ Sort the scores;
- ÷ Outputs: [*FS_Array* - array of selection scores; *FSD_Array* - array of distinct selection scores; *FSC_Array* - occurrences of every distinct selection score].

***Proportional selection strategy*** comes from (*N_Sel* - number of selections to do):

- ÷ Set *Selected_Genotypes_Array* to *Empty*;
- ÷ For every selection from 1 to *N_Sel* do

- Compute the sum of unselected genotypes scores to *FS_Sum*;
- Random (uniform distribution) generate a number *FS_Freq* between 0 and *FS_Sum* (inclusive);
- Find first index *Group* from *FSD_Array* for which *FS_Freq* ≤ $\Sigma_{i\leq\text{Group}}$*FSD_Array[i]*FSC_Array[i]*;
- Random (uniform distribution) generate a number *FSD_Next* between 1 and *FSC_Array[i]*;
- Add to *Selected_Genotypes_Array* the *FSD_Next* value (not selected yet) of *FSD_Array[Group]* from *FS_Array* and decrease *FSC_Array[Group]* with one.

***Deterministic selection strategy*** comes from:

÷ Set *Selected_Genotypes_Array* to *Empty*, *Already_Selected* to 0, *Group* to sample size;

÷ While *Already_Selected* + *FSC_Array[Group]<=N_Sel* put the indices from FS_Array equal with *FSD_Array[Group]* into *Selected_Genotypes_Array* and decrease *Group* with one if is possible or increase *Group* otherwise;

÷ While *Already_Selected<=N_Sel* (full groups are exhausted; only a part of the group will be selected here);
  - Random (uniform distribution) generate a number *FSD_Next* between 1 and *FSC_Array[i]*;
  - Add to *Selected_Genotypes_Array* the *FSD_Next* value (not selected yet) of *FSD_Array[Group]* from *FS_Array* and decrease *FSC_Array[Group]* with one.

***Tournament selection strategy*** comes from:

÷ Let *N_Gen* to be the number of genotypes from sample;

÷ Random (uniform distribution) generate a permutation of {1..*N_Gen*} into *Selected_Genotypes_Array*;

÷ For every *i_Sel* from 2 to *N_Sel* (first N_Sel competes in tournament)
  - If *FS_Array[i_Sel]≤FS_Array[i_Sel-1]* then
    - If *FS_Array[i_Sel]=FS_Array[i_Sel-1]* then if random selection between 0 and 1 gives 0 then continue (for iteration);
    - Swift in *FS_Array* the values from *i_Sel* and *i_Sel-1*;

÷ If *N_sel<N_Gen* then (last selected did not participate in tournament and still are elements with which to compete in sample)
  - Random (uniform distribution) generate a number *i_Sel* between *N_Sel+1* and *N_Gen*;
  - If *FS_Array[N_Sel]≤FS_Array[i_Sel]* then
    - If *FS_Array[N_Sel]=FS_Array[i_Sel]* then if random selection between 0 and 1 gives 0 then stop (tournament completed);
    - Swift in *FS_Array* the values from *i_Sel* and *N_Sel*.

On ***survival scores VS*** (Table 6) are made same calculations as for selection scores (FS). ***Proportional survival strategy*** same procedure (on VS) were applied as for proportional selection (on FS). ***Deterministic survival strategy*** same procedure (on VS) were applied as for deterministic selection (on FS). ***Tournament survival strategy*** same procedure (on VS) were applied as for tournament selection (on FS).

***Ranks*** are obtained from:

÷ Ranking the values as in Spearman correlation coefficient algorithm [52,53];

÷ Multiplies by two, decrease first one from all and add one (to be integers starting from 1);

÷ Rank sorting: QuickSort [54] algorithm (requiring 2·[$\log_2(n)$] crossovers, as in Figure 6).

The ***evolutionary program*** (the program which implements the genetic algorithm) was build in order to be able to work with any family of molecular descriptors, and was parameterized through a series of configuration files.

| Positions | 0 | 1 | 2 | 3 | 4 | 5 | 6 | 7 | 8 | 9 |
|---|---|---|---|---|---|---|---|---|---|---|
| Initial array | 3 | 1 | 4 | 1 | 5 | **9** | 2 | 6 | **5** | 5 |
| Crossover | | | | | | | | | | |
| Step 1 | **3** | 1 | 4 | **1** | 5 | **5** | 2 | 6 | **9** | 5 |
| Crossover | | | | | | | | | | |
| Step 2 | **1** | 1 | 4 | **3** | **5** | 5 | **2** | 6 | 9 | 5 |
| Crossover | | | | | | | | | | |
| Step 3 | 1 | 1 | **4** | 3 | **2** | 5 | **5** | 6 | 9 | 5 |
| Crossover | | | | | | | | | | |
| Step 4 | 1 | 1 | **2** | 3 | **4** | 5 | 5 | 6 | **9** | **5** |
| Crossover | | | | | | | | | | |
| Step 5 | 1 | 1 | 2 | 3 | 4 | 5 | 5 | **6** | **5** | **9** |
| Crossover | | | | | | | | | | |
| Step 6 | 1 | 1 | 2 | 3 | 4 | 5 | 5 | **5** | **6** | 9 |
| Sorted array | 1 | 1 | 2 | 3 | 4 | 5 | 5 | 5 | 6 | 9 |

**Figure 6.** A QuickSort example

The program uses a configuration file for the connectivity with the database storing the molecular descriptors, as in Figure 7 below.

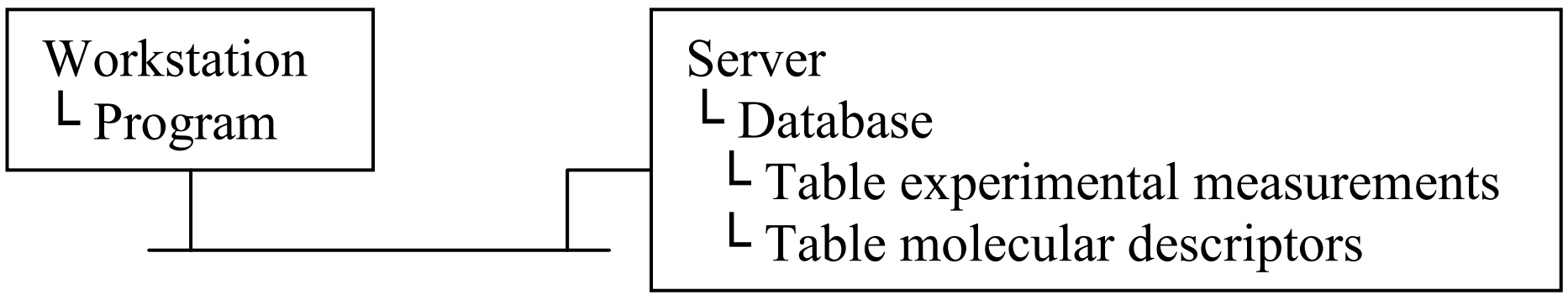

**Figure 7.** Design of client-server connectivity topology

A connection to a database on a server requires security protocols. The c_galg.cfg configuration file specifies the required information (Figure 8a). The next configuration file (c_galg.cgt) contains the definition of the genetic topology of the descriptors family. Figure 8b shows the content of the c_galg.cgt file for MDF genetic topology (Table 1), as it was run the program, results being described in following section.

The values of a series of parameters defining the evolution of the genetic algorithm were stored in a third configuration file, c_galg.cga (Figure 9).

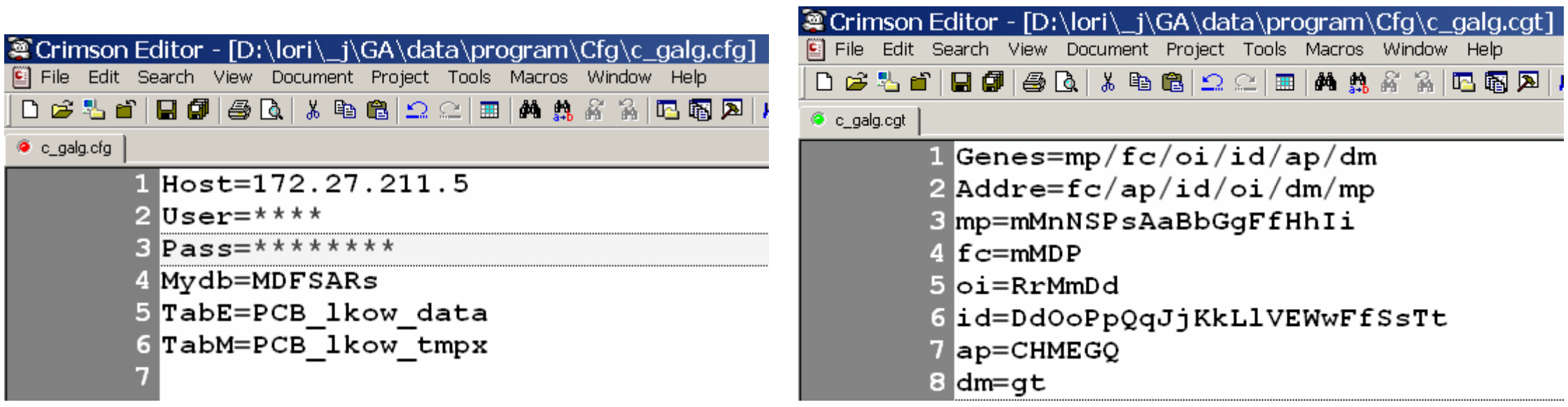

(a) (b)

**Figure 8.** Configuration: database connectivity (left) and genetic topology for MDF (right)

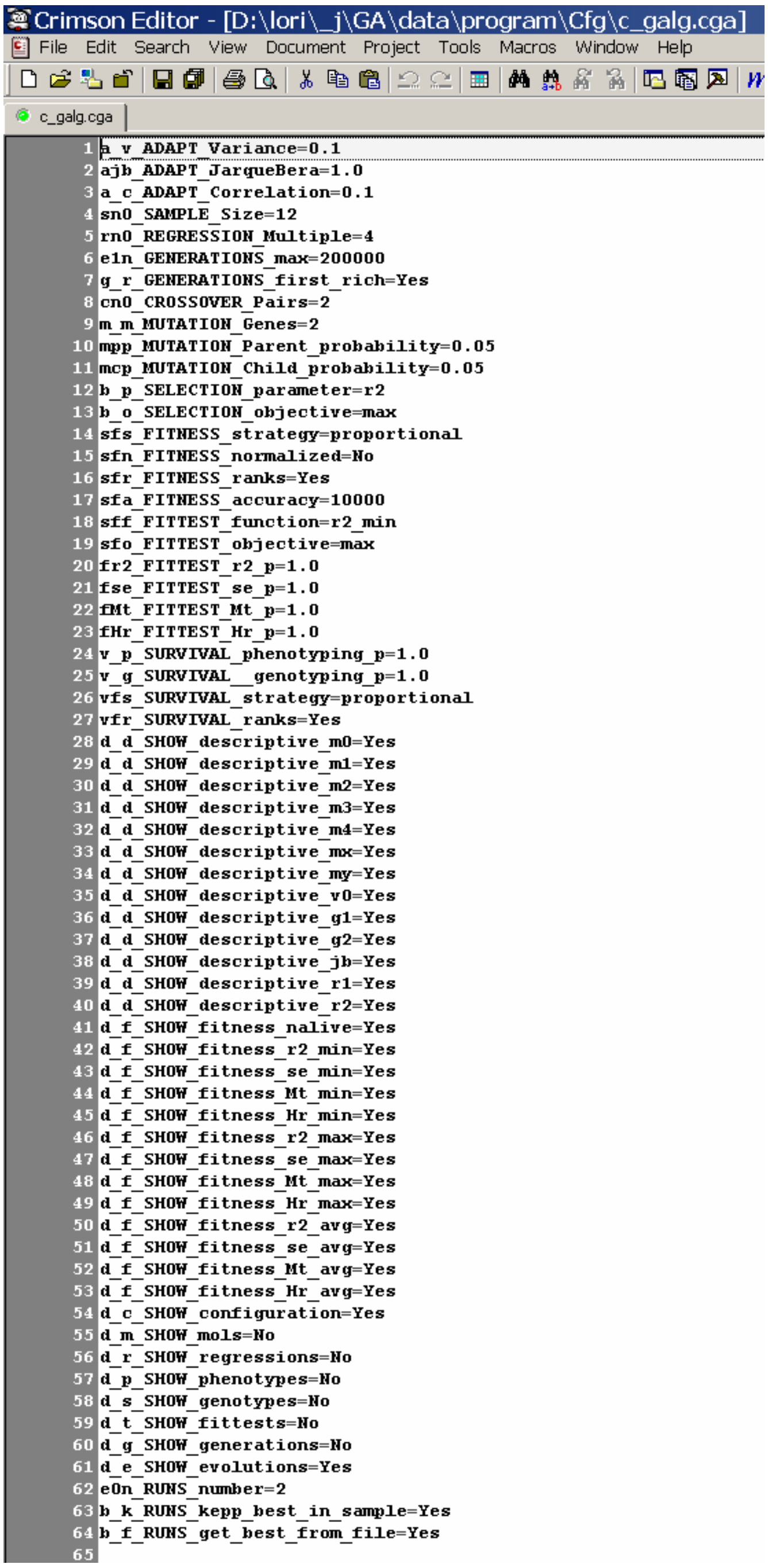

**Figure 9.** Genetic algorithm evolution configuration file

## 4. Results and discussion

In order to test the program, an experiment of selection and survival strategy was designed (Table 7) and were run on five dual core processor based machines. In order to avoid the overwriting of the files from a program to another, a random number were added automatically by the program to the name of the output file, as is given in Table 7. The program was run for 20000 generations on 206 PCBs by using logarithm of octanol/water partition coefficient data, already available in MDF database from previous investigation ([48], http://l.academicdirect.org/Chemistry/SARs/MDF_SARs/). Figure 10 contains the screen prints of the obtained results, the files being available upon request.

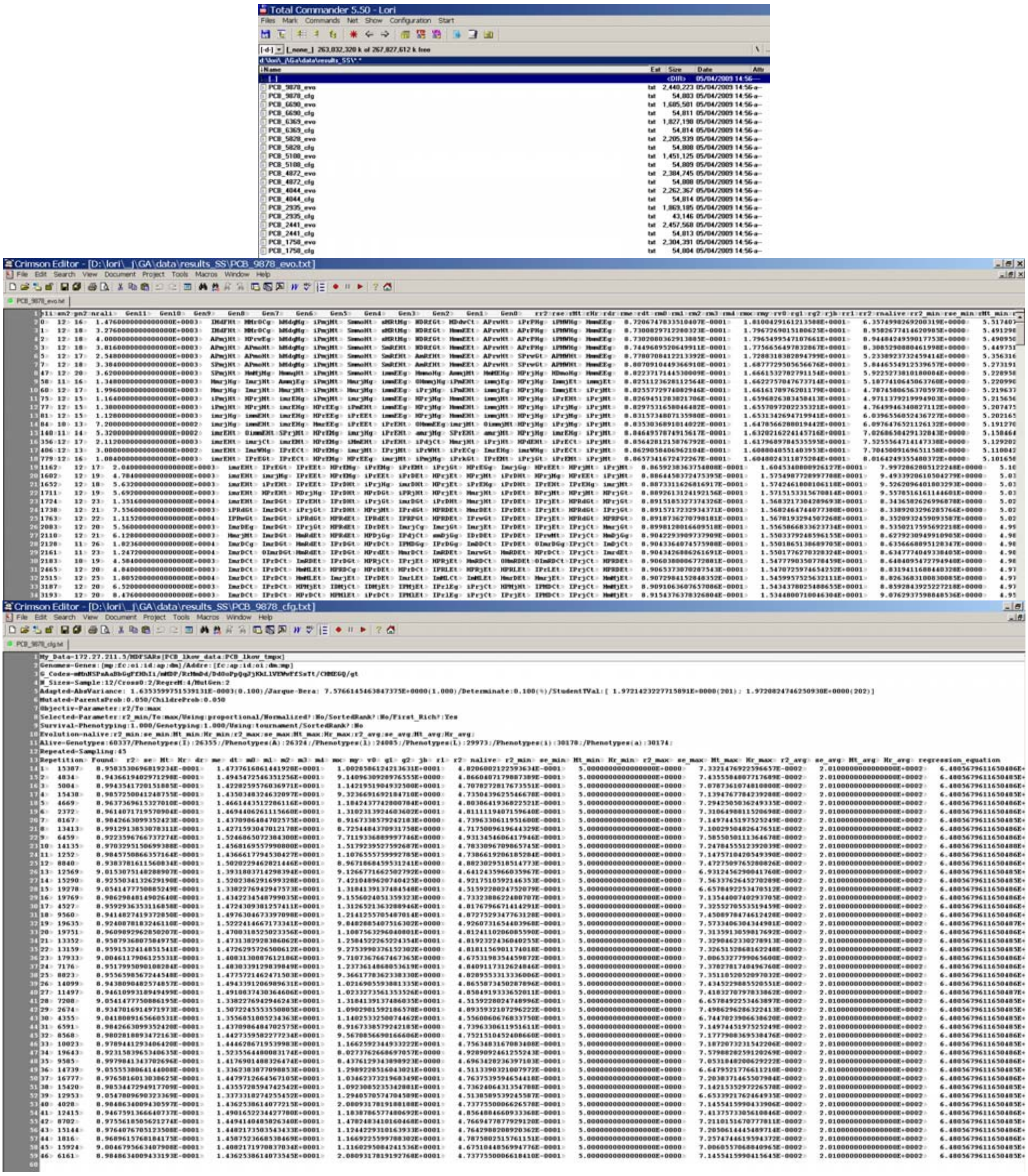

**Figure 10.** Output files on PCB (top: all files; middle: 9878_evo file; bottom: 9878_cfg)

**Table 7.** Experimental design: selection and survival strategies

| Selection \ Survival | Proportional (P) | Deterministic (D) | Tournament (T) |
|---|---|---|---|
| Proportional (P) | P:P (4044) | P:D (2441) | P:T (9878) |
| Deterministic (D) | D:P (5108) | D:D (6369) | D:T (6690) |
| Tournament (T) | T:P (5828) | T:D (4872) | T:T (1758) |

**Table 8.** Most frequent genotypes found in the generations which lead to evolution (improving of the objective function) from 46 independent runs

Proportional selection strategy

| VS | Gen | Num | Occ | Par |
|---|---|---|---|---|
| P | T23 | 13 | 406 | 389 |
| | mMdlHg | 1 | 46 | 43 |
| | MDMKHt | 1 | 40 | 39 |
| | nDRLHt | 1 | 40 | 39 |
| | iPDKCg | 1 | 39 | 39 |
| | ADDJCg | 1 | 35 | 35 |
| | mDdjGg | 1 | 31 | 30 |
| | bDDDGg | 1 | 28 | 19 |
| | bDDJCg | 1 | 27 | 27 |
| | sDdLHg | 1 | 25 | 25 |
| | BDDDGg | 1 | 24 | 22 |
| | bDMLEg | 1 | 24 | 24 |
| | bDMLGg | 1 | 24 | 24 |
| | MMDPMt | 1 | 23 | 23 |
| | Tot | 6760 | 16788 | 15902 |
| D | T23 | 13 | 378 | 371 |
| | iPMDHg | 1 | 39 | 37 |
| | bPRjCg | 1 | 38 | 38 |
| | IPMDEg | 1 | 37 | 36 |
| | mMdoHt | 1 | 30 | 29 |
| | IPRKCg | 1 | 29 | 29 |
| | MDRLHt | 1 | 29 | 29 |
| | MMdlHg | 1 | 29 | 29 |
| | MDmWHg | 1 | 26 | 26 |
| | BPRjCg | 1 | 26 | 25 |
| | NDRlHt | 1 | 25 | 25 |
| | iPMDCg | 1 | 24 | 23 |
| | bmrVCt | 1 | 23 | 23 |
| | IPMDCg | 1 | 23 | 22 |
| | Tot | 8070 | 18240 | 17797 |
| T | T23 | 6 | 214 | 207 |
| | MMdlHg | 1 | 47 | 47 |
| | mMdlHg | 1 | 46 | 43 |
| | sPDLEg | 1 | 38 | 38 |
| | AMdwGg | 1 | 29 | 29 |
| | IPMDHg | 1 | 29 | 27 |
| | mMdqGt | 1 | 25 | 23 |
| | Tot | 7466 | 16599 | 15739 |

Deterministic selection strategy

| VS | Gen | Num | Occ | Par |
|---|---|---|---|---|
| P | T23 | 3 | 89 | 72 |
| | MDRLHt | 1 | 31 | 31 |
| | ImrWCg | 1 | 30 | 19 |
| | ImrWHg | 1 | 28 | 22 |
| | Total | 3922 | 10764 | 9742 |
| D | T23 | 32 | 893 | 893 |
| | gmdKHg | 1 | 48 | 48 |
| | iPDDGg | 1 | 43 | 43 |
| | bmRkHg | 1 | 37 | 37 |
| | gMdEQg | 1 | 34 | 34 |
| | sDRDGg | 1 | 34 | 34 |
| | HDmLQt | 1 | 33 | 33 |
| | MDMKHt | 1 | 33 | 33 |
| | mMdLMt | 1 | 30 | 30 |
| | MMmwCg | 1 | 29 | 29 |
| | bmdFEt | 1 | 29 | 29 |
| | hDDJCg | 1 | 27 | 27 |
| | hDDpCg | 1 | 27 | 27 |
| | hPmEMg | 1 | 27 | 27 |
| | sPmJMt | 1 | 27 | 27 |
| | NmdlQg | 1 | 26 | 26 |
| | SMMFEg | 1 | 26 | 26 |
| | bMddEg | 1 | 26 | 26 |
| | sPRDHt | 1 | 26 | 26 |
| | BDrsGt | 1 | 25 | 25 |
| | hDMKEg | 1 | 25 | 25 |
| | smdoQg | 1 | 25 | 25 |
| | AMMpHt | 1 | 24 | 24 |
| | GPmVCg | 1 | 24 | 24 |
| | SMMjEt | 1 | 24 | 24 |
| | BPMkHg | 1 | 23 | 23 |
| | GmmlQt | 1 | 23 | 23 |
| | bPmjMg | 1 | 23 | 23 |
| | hDDDHg | 1 | 23 | 23 |
| | hMdWGt | 1 | 23 | 23 |
| | hPmSEg | 1 | 23 | 23 |
| | hmddCt | 1 | 23 | 23 |
| | imMtGg | 1 | 23 | 23 |
| | Tot | 4385 | 13560 | 13316 |
| T | T23 | 5 | 152 | 152 |
| | NDRkHt | 1 | 37 | 37 |
| | sDDEMg | 1 | 30 | 30 |
| | hMrkGg | 1 | 29 | 29 |
| | MDDKHt | 1 | 28 | 28 |
| | sMrLCg | 1 | 28 | 28 |
| | Tot | 4965 | 12504 | 11572 |

Tournament selection strategy

| VS | Gen | Num | Occ | Par |
|---|---|---|---|---|
| P | T23 | 13 | 419 | 405 |
| | sPDJEg | 1 | 64 | 64 |
| | mMdlHg | 1 | 44 | 42 |
| | MMdlHg | 1 | 40 | 40 |
| | MDdjEg | 1 | 32 | 30 |
| | sDMDM | 1 | 29 | 28 |
| | g | 1 | 29 | 23 |
| | mMdqGt | 1 | 28 | 28 |
| | sDDKCg | 1 | 28 | 28 |
| | sPDLEg | 1 | 27 | 27 |
| | aDDKEg | 1 | 26 | 26 |
| | sDRKCg | 1 | 25 | 22 |
| | sPRKGg | 1 | 24 | 24 |
| | sDMLGg | 1 | 23 | 23 |
| | MDRLHt | | | |
| | Tot | 6537 | 16368 | 15317 |
| D | T23 | 21 | 714 | 687 |
| | MDRLHt | 1 | 88 | 87 |
| | IPMJCg | 1 | 46 | 45 |
| | IPMDEg | 1 | 42 | 38 |
| | sDRJEg | 1 | 41 | 39 |
| | iPMKCg | 1 | 36 | 36 |
| | iPDJCg | 1 | 35 | 33 |
| | sPDLEg | 1 | 34 | 34 |
| | mDRlHt | 1 | 33 | 33 |
| | nDRLHt | 1 | 32 | 31 |
| | sDMLCg | 1 | 31 | 29 |
| | iPDDGg | 1 | 31 | 28 |
| | iPDDEg | 1 | 29 | 27 |
| | mDRkHt | 1 | 28 | 28 |
| | IPRKCg | 1 | 27 | 26 |
| | IPDJCg | 1 | 27 | 25 |
| | iPDKCg | 1 | 27 | 25 |
| | bPmkEt | 1 | 26 | 26 |
| | sDDJEg | 1 | 26 | 26 |
| | MDDKHt | 1 | 26 | 22 |
| | IPDKCg | 1 | 25 | 25 |
| | sDDLHg | 1 | 24 | 24 |
| | Tot | 7964 | 17700 | 17331 |
| T | T23 | 8 | 217 | 213 |
| | IDRwHt | 1 | 34 | 34 |
| | mMdlHg | 1 | 28 | 28 |
| | nMRSEt | 1 | 28 | 27 |
| | mPRDHt | 1 | 27 | 26 |
| | MDRLHt | 1 | 26 | 26 |
| | smmLCt | 1 | 26 | 24 |
| | AMDEQt | 1 | 24 | 24 |
| | IDRwGt | 1 | 24 | 24 |
| | Tot | 7529 | 17100 | 16151 |

VS: Survival strategy; P: Proportional; T: Tournament; D: Deterministic;
Gen: Genotypes; Num: Number (of distinct genotypes); Occ: Occurrences (of the genotypes);
Par: Participations in valid regressions (of the genotypes);
T23: Top of genotypes with 23 or more occurrences; Tot: all occurrences.

The frequency of the genotypes without linearization operator (last gene of the MDF family) may be a measure of the adaptation and in same time a measure for variability of the genetic material produced by the selection and survival strategy. In order to avoid the bias of the chance, 46 runs were done for every pair of selection and survival strategies. Above table (Table 8) contains this information resulted after descriptive processing of *_evo.txt files.

The chi-square statistic [55-57] may be used to test the homogeneity of the genotype populations obtained via different selection and survival strategies.

The following tables (Table 9 - Table 14) check homogeneity hypotheses regarding the number of genotypes found in the evolution leading generations. The tables contain the observed numbers and into parentheses the expected numbers under homogeneity hypothesis.

**Table 9.** Are homogenous the populations of number of distinct genotypes when we draw observations from different selection and survival strategies?

| $\chi^2$ | P: Obs. (Exp.) | T: Obs. (Exp.) | D: Obs. (Exp.) | Σ |
|---|---|---|---|---|
| P | 6760 (6665) | 7466 (7726) | 8070 (7904) | 22296 |
| T | 6537 (6586) | 7529 (7634) | 7964 (7810) | 22030 |
| D | 3922 (3968) | 4965 (4599) | 4385 (4705) | 13272 |
| Σ | 17219 | 19960 | 20419 | 57598 |

| Unexplained square error | Probability from Chi Square distribution | Answer |
|---|---|---|
| $X^2(P,\cdot) = 13.6$ | $p_{\chi 2}(x^2 > X^2,2) = 1‰$ | No |
| $X^2(T,\cdot) = 4.85$ | $p_{\chi 2}(x^2 > X^2,2) = 9\%$ | - |
| $X^2(D,\cdot) = 51.4$ | $p_{\chi 2}(x^2 > X^2,2) = 7\cdot 10^{-12}$ | No |
| $X^2(\cdot,P) = 2.25$ | $p_{\chi 2}(x^2 > X^2,2) = 32\%$ | - |
| $X^2(\cdot,T) = 39.3$ | $p_{\chi 2}(x^2 > X^2,2) = 3\cdot 10^{-9}$ | No |
| $X^2(\cdot,D) = 28.3$ | $p_{\chi 2}(x^2 > X^2,2) = 7\cdot 10^{-7}$ | No |
| $X^2(\cdot,\cdot) = 69.9$ | $p_{\chi 2}(x^2 > X^2,4) = 2\cdot 10^{-14}$ | No |

**Table 10.** Are homogenous the populations of total number of genotypes when we draw observations from different selection and survival strategies?

| $\chi^2$ | P | T | D | Σ |
|---|---|---|---|---|
| P | 16788 (16240) | 16599 (17084) | 18240 (18303) | 51627 |
| T | 16368 (16095) | 17100 (16932) | 17700 (18140) | 51168 |
| D | 10764 (11585) | 12504 (12187) | 13560 (13056) | 36828 |
| Σ | 43920 | 46203 | 49500 | 139623 |

| Unexplained square error | Probability from Chi Square distribution | Answer |
|---|---|---|
| $X^2(P,\cdot) = 32.5$ | $p_{\chi 2}(x^2 > X^2,2) = 9\cdot 10^{-8}$ | No |
| $X^2(T,\cdot) = 17.0$ | $p_{\chi 2}(x^2 > X^2,2) = 2\cdot 10^{-4}$ | No |
| $X^2(D,\cdot) = 85.9$ | $p_{\chi 2}(x^2 > X^2,2) = 2\cdot 10^{-19}$ | No |
| $X^2(\cdot,P) = 81.3$ | $p_{\chi 2}(x^2 > X^2,2) = 2\cdot 10^{-18}$ | No |
| $X^2(\cdot,T) = 23.7$ | $p_{\chi 2}(x^2 > X^2,2) = 7\cdot 10^{-6}$ | No |
| $X^2(\cdot,D) = 30.3$ | $p_{\chi 2}(x^2 > X^2,2) = 3\cdot 10^{-7}$ | No |
| $X^2(\cdot,\cdot) = 135$ | $p_{\chi 2}(x^2 > X^2,4) = 3\cdot 10^{-28}$ | No |

**Table 11.** Are homogenous the populations of genotypes which provide valid regressions when we draw observations from different selection and survival strategies?

| $\chi^2$ | P | T | D | Σ |
|---|---|---|---|---|
| P | 15902 (15241) | 15739 (16172) | 17797 (18025) | 49438 |
| T | 15317 (15044) | 16151 (15963) | 17331 (17792) | 48799 |
| D | 9742 (10676) | 11572 (11328) | 13316 (12626) | 34630 |
| Σ | 40961 | 43462 | 48444 | 132867 |

| Unexplained square error | Probability from Chi Square distribution | Answer |
|---|---|---|
| $X^2(P,\cdot) = 43.1$ | $p_{\chi 2}(x^2 > X^2,2) = 4\cdot 10^{-10}$ | No |
| $X^2(T,\cdot) = 19.1$ | $p_{\chi 2}(x^2 > X^2,2) = 7\cdot 10^{-5}$ | No |
| $X^2(D,\cdot) = 125$ | $p_{\chi 2}(x^2 > X^2,2) = 8\cdot 10^{-28}$ | No |
| $X^2(\cdot,P) = 115$ | $p_{\chi 2}(x^2 > X^2,2) = 9\cdot 10^{-26}$ | No |
| $X^2(\cdot,T) = 19.1$ | $p_{\chi 2}(x^2 > X^2,2) = 7\cdot 10^{-5}$ | No |
| $X^2(\cdot,D) = 52.5$ | $p_{\chi 2}(x^2 > X^2,2) = 4\cdot 10^{-12}$ | No |
| $X^2(\cdot,\cdot) = 187$ | $p_{\chi 2}(x^2 > X^2,4) = 2\cdot 10^{-39}$ | No |

**Table 12.** Are homogenous the populations of number of distinct genotypes from Top 23 when we draw observations from different selection and survival strategies?

| $\chi^2$ | P | T | D | Σ |
|---|---|---|---|---|
| P | 13 (8) | 6 (5) | 13 (19) | 32 |
| T | 13 (11) | 8 (7) | 21 (24) | 42 |
| D | 3 (10) | 5 (7) | 32 (23) | 40 |
| Σ | 29 | 19 | 66 | 114 |

| Unexplained square error | Probability from Chi Square distribution | Answer |
|---|---|---|
| $X^2(P,\cdot) = 5.22$ | $p_{\chi 2}(x^2 > X^2,2) = 7.4\%$ | - |
| $X^2(T,\cdot) = 0.88$ | $p_{\chi 2}(x^2 > X^2,2) = 64\%$ | - |
| $X^2(D,\cdot) = 8.99$ | $p_{\chi 2}(x^2 > X^2,2) = 1.1\%$ | No |
| $X^2(\cdot,P) = 8.39$ | $p_{\chi 2}(x^2 > X^2,2) = 1.5\%$ | No |
| $X^2(\cdot,T) = 0.91$ | $p_{\chi 2}(x^2 > X^2,2) = 63\%$ | - |
| $X^2(\cdot,D) = 5.79$ | $p_{\chi 2}(x^2 > X^2,2) = 5.5\%$ | - |
| $X^2(\cdot,\cdot) = 15.1$ | $p_{\chi 2}(x^2 > X^2,4) = 4.5‰$ | No |

**Table 13.** Are homogenous the populations of total number of genotypes from Top 23 when we draw observations from different selection and survival strategies?

| $\chi^2$ | P | T | D | Σ |
|---|---|---|---|---|
| P | 406 (262) | 214 (167) | 378 (569) | 998 |
| T | 419 (354) | 217 (226) | 714 (770) | 1350 |
| D | 89 (298) | 152 (190) | 893 (646) | 1134 |
| Σ | 914 | 583 | 1985 | 3482 |

| Unexplained square error | Probability from Chi Square distribution | Answer |
|---|---|---|
| $X^2(P,\cdot) = 156$ | $p_{\chi 2}(x^2 > X^2,2) = 10^{-34}$ | No |
| $X^2(T,\cdot) = 16.4$ | $p_{\chi 2}(x^2 > X^2,2) = 0.3‰$ | No |
| $X^2(D,\cdot) = 249$ | $p_{\chi 2}(x^2 > X^2,2) = 10^{-54}$ | No |
| $X^2(\cdot,P) = 238$ | $p_{\chi 2}(x^2 > X^2,2) = 2\cdot 10^{-52}$ | No |
| $X^2(\cdot,T) = 21.2$ | $p_{\chi 2}(x^2 > X^2,2) = 3\cdot 10^{-5}$ | No |
| $X^2(\cdot,D) = 163$ | $p_{\chi 2}(x^2 > X^2,2) = 5\cdot 10^{-36}$ | No |
| $X^2(\cdot,\cdot) = 421$ | $p_{\chi 2}(x^2 > X^2,4) = 6\cdot 10^{-90}$ | No |

**Table 14.** Are homogenous the populations of genotypes from Top 23 which provide valid regressions when we draw observations from different selection and survival strategies?

| $\chi^2$ | P | T | D | Σ |
|---|---|---|---|---|
| P | 389 (247) | 207 (163) | 371 (557) | 967 |
| T | 405 (333) | 213 (220) | 687 (751) | 1305 |
| D | 72 (285) | 152 (189) | 893 (643) | 1117 |
| Σ | 866 | 572 | 1951 | 3389 |

| Unexplained square error | Probability from Chi Square distribution | Answer |
|---|---|---|
| $X^2(P,\cdot) = 156$ | $p_{\chi 2}(x^2 > X^2,2) = 2\cdot 10^{-34}$ | No |
| $X^2(T,\cdot) = 21.2$ | $p_{\chi 2}(x^2 > X^2,2) = 2\cdot 10^{-5}$ | No |
| $X^2(D,\cdot) = 264$ | $p_{\chi 2}(x^2 > X^2,2) = 6\cdot 10^{-58}$ | No |
| $X^2(\cdot,P) = 256$ | $p_{\chi 2}(x^2 > X^2,2) = 2\cdot 10^{-56}$ | No |
| $X^2(\cdot,T) = 19.3$ | $p_{\chi 2}(x^2 > X^2,2) = 6\cdot 10^{-5}$ | No |
| $X^2(\cdot,D) = 165$ | $p_{\chi 2}(x^2 > X^2,2) = 2\cdot 10^{-36}$ | No |
| $X^2(\cdot,\cdot) = 441$ | $p_{\chi 2}(x^2 > X^2,4) = 5\cdot 10^{-94}$ | No |

## 5. Conclusions

A genetic algorithm for multiple linear regressions with families of descriptors for structure-activity relationships was implemented and tested.

From 46 runs on a set of 206 PCBs relating their structure with their logarithm of octanol/water partition coefficient by using an experimental design meant to compare the obtained results through different selection and survival strategies during evolution shows that different selection and survival strategies create different partitions on the entire population of all possible genotypes.